\documentclass[a4paper,twoside,12pt]{article}

\usepackage[utf8]{inputenc}
\usepackage[T1]{fontenc}
\usepackage[a4paper,twoside,top=2.5cm,bottom=4cm,inner=3cm,outer=2cm,footskip=2cm]{geometry}
\usepackage{setspace}
\usepackage{amsmath}
\usepackage{amssymb}
\usepackage{amsthm}
\usepackage[unicode,bookmarksopen]{hyperref}
\usepackage{pbox}
\usepackage{cite}
\usepackage{verbatim}
\usepackage{moreverb}

\newcommand{\attach}[1]{\tiny\verbatimtabinput{#1}\normalsize }

\begin{document}

\title{An implementation of the relational k-means algorithm}
\author{Balázs Szalkai \\{\small Eötvös Loránd University, Budapest, Hungary}}
\maketitle

\begin{abstract}
A $C\sharp$ implementation of a generalized k-means variant called {\em relational k-means} is described here. Relational k-means is a generalization of the well-known k-means clustering method which works for non-Euclidean scenarios as well. The input is an arbitrary distance matrix, as opposed to the traditional k-means method, where the clustered objects need to be identified with vectors.
\end{abstract}

\section{Introduction}

Suppose that we are given a set of data points $p_1, ... p_n \in \mathbb{R}^d$ and a number of clusters $N$. We would like assort these data points ``optimally'' into $N$ clusters. For each point $p_i$ let $z_i$ denote the center of gravity of the cluster $p_i$ is assigned to. We call the $z_i$ vectors {\em centroids}. The standard k-means method \cite{kmeans} attempts to produce such a clustering of the data points that the sum of squared centroid distances $\sum_{i = 1}^{n} ||p_i - z_i||^2$ is minimized.

The main difficulty of this method is that it requires the data points to be the elements of a Euclidean space, since we need to average the data points somehow. A generalization of k-means called {\em relational k-means} has been proposed \cite{2013arXiv1303.6001S} to address this issue. This generalized k-means variant does not require the data points to be vectors.

Instead, the data points $p_1, ... p_n$ can form an abstract set $S$ with a completely arbitrary distance function $f : S \times S \to [0, \infty)$. We only require that $f$ is symmetrical, and $f(p_i, p_i) = 0$ for all $p_i \in S$. Note that $f$ need not even satisfy the triangle inequality.

\section{Relational k-means}

First, we provide a brief outline of the algorithm. Let $A \in \mathbb{R}^{n \times n}$ be the squared distance matrix. That is, $A_{ij} = f(p_i, p_j)^2$. The algorithm starts with some initial clustering and improves it by repeatedly performing an iteration step. The algorithm stops if the last iteration did not decrease the {\em value} of the clustering (defined below). Of course, if the iteration increased the value of the clustering, the algorithm reverts the last iteration.

Now we describe the algorithm in detail. At any time during execution, let $S_1, ... S_N \subset S$ denote the clusters. For each data point $p_i$, let $\ell(p_i)$ denote the index of the cluster $p_i$ is assigned to. Let $e_i \in \mathbb{R}^n$ denote the $i$th standard basis vector, and, for $i \in \{1, ... n\}$ and $j \in \{1, ... N\}$ let $v_{ij} := \frac{1}{|S_j|}\sum_{k \in S_j} e_k - e_i$. Let us call the quantity $q_{ij} := -\frac{1}{2}v_{ij}^\top A v_{ij}$ the {\em squared centroid distance} corresponding to the point $p_i$ and the cluster $S_j$.

In \cite{2013arXiv1303.6001S} it is observed that, if the distance function is derived from a Euclidean representation of the data points, then $q_{ij}$ equals to the squared distance of $p_i$ and the center of gravity of $S_j$. Thus $q_{ij}$ is indeed an extension of the classical notion of squared centroid distances.

Define the {\em value} of a clustering as $\sum_{i = 1}^n q_{i\ell{i}}$. We say that a clustering is {\em better} than another one if and only if its value is less than the value of the other clustering.

The relational k-means algorithm takes an initial clustering (e.g. a random one), and improves it through repeated applications of an iteration step which reclusters the data points. The iteration step simply reassigns each data point to the cluster which minimized the squared centroid distance in the previous clustering. If the value of clustering does not decrease through the reassignment, then the reassignment is undone and the algorithm stops.

In the non-Euclidean case there might be scenarios when an iteration actually {\em worsens} the clustering. Should this peculiar behavior be undesirable, it can be avoided by ``stretching'' the distance matrix and thus making it Euclidean.

Stretching means replacing $A$ with $A + \beta (J-I)$, where $J$ is the matrix whose entries are all 1's, $I$ is the identity matrix, and $\beta \geq 0$ is the smallest real number for which $A + \beta (J-I)$ is a Euclidean squared distance matrix, i.e. it equals to the squared distance matrix of some $n$ vectors. It can be easily deducted that such a $\beta$ exists. This method is called $\beta$-spread transformation (see \cite{Hathaway1994429}).

The algorithm is thus as follows:

\begin{itemize}
\item Start with some clustering, e.g. a random one
\item Calculate the value of the current clustering and store it in $V_1$
\item For each $p_i$ data point, calculate and store the squared centroid distances $q_{i1}, ... q_{iN}$
\item Reassign each $p_i$ data point to the cluster that yielded the smallest squared centroid distance in the previous step
\item Calculate the value of the current clustering and store it in $V_2$
\item If $V_2 \geq V_1$, then undo the previous reassignment and stop
\item Go to line number 2
\end{itemize}

\section{Time complexity}

The algorithm is clearly finite because it gradually decreases the value of the current clustering and the number of different clusterings is finite. Each iteration step can easily be implemented in $\mathcal{O}(n^3)$ time: for each data point, we need to calculate $N$ quadratic forms, which can be done in $n \sum_{j=1}^N |S_j|^2 \leq n^3$ time. This is unfortunately too slow for practical applications.

However, this can be improved down to $\mathcal{O}(n^2)$. The squared centroid distances can be transformed as follows (using $A_{ii} = 0$):

\[
q_{ij} = -\frac{1}{2}\left(\frac{1}{|S_j|}\sum_{k \in S_j} e_k - e_i\right)^\top A \left(\frac{1}{|S_j|}\sum_{k \in S_j} e_k - e_i\right) = -\frac{1}{2|S_j|^2} \sum_{a, b \in S_j} A_{ab} + \frac{1}{|S_j|} \sum_{k \in S_j} A_{ik}.
\]

Here the first summand is independent of $i$ and thus needs to be calculated for each cluster only once per iteration. On the other hand, the second summand can be calculated in $\mathcal{O}(|S_j|)$ time. To sum up, the amount of arithmetic operations per iteration is at most constant times $\sum_{j = 1}^N |S_j|^2 + n \sum_{j = 1}^N |S_j| \leq 2 n^2$.

The current implementation makes several attempts to find a better clustering. In each attempt, the full relational k-means algorithm is run, starting from a new random clustering. Every attempt has the possibility to produce a clustering which is better than the best one among the previous attempts. If the sofar best clustering has not been improved in the last $K$ attempts (where $K$ is a parameter), then it is assumed that the clustering which is currently the best is not too far from the optimum, and the program execution stops.

Attempts do not depend on each other's result and do not modify shared resources (apart from a shared random generator). Our implementation uses \texttt{Parallel.For} for launching multiple attempts at once. The number of parallel threads can be customized via a command-line switch, and by default equals to the number of logical processors. This results in near 100\% CPU utilization. Launching less than $C$ threads allows leaving some CPU time for other processes.

\section{Test results and conclusion}

We implemented the above algorithm in $C\sharp$ and run the program on a set of >1000 proteins with a Levenshtein-like distance function. The value of $K$ (maximum allowed number of failed attempts, i.e. the ``bad luck streak'') was 20, and the value of $N$ (number of clusters) was 10. The testing was done on an average dual core laptop computer and finished in 30..60 seconds. This proves that relational k-means can be implemented in a way efficient enough to be applied to real-world datasets.

Since the program is almost fully parallelized, we could expect it to finish in <8 seconds for the same dataset on a 16-core machine. Note that the total runtime is proportional to the number of attempts made, which is highly variable due to the random nature of the algorithm.

A C++ implementation could further reduce execution time. According to our estimate, it could make the program cca. twice as fast.

\section{Attachments}

\subsection{Program source code in $C\sharp$}
\attach{Program.cs}
\newpage

\subsection{Example input}

The first $n$ lines of the input contain the names of the objects which need to be clustered. Then a line containing two slashes follows. After that, a matrix containing the pairwise distances is listed in semicolon-separated CSV format. Fractional distances must be input with the dot character (\texttt{.}) as decimal separator.

\attach{testinput.txt}

\bibliography{generalized-kmeans}
\bibliographystyle{plain}

\end{document}